\documentclass[conference]{IEEEtran}
\IEEEoverridecommandlockouts
\usepackage{cite}
\usepackage[caption=false,font=normalsize,labelfont=rm,textfont=rm]{subfig}
\usepackage{longtable}
\usepackage{amsmath}
\usepackage{amssymb}
\usepackage{multirow}
\usepackage{verbatim}
\usepackage{graphicx}
\usepackage{utfsym}
\def\BibTeX{{\rm B\kern-.05em{\sc i\kern-.025em b}\kern-.08em
    T\kern-.1667em\lower.7ex\hbox{E}\kern-.125emX}}
\begin{document}
\title{FreCT: Frequency-augmented Convolutional Transformer for Robust Time Series Anomaly Detection}
%
%

\author{

\IEEEauthorblockN{
Wenxin Zhang \IEEEauthorrefmark{1}, 
Ding Xu \IEEEauthorrefmark{2},
Guangzhen Yao \IEEEauthorrefmark{3},
Xiaojian Lin \IEEEauthorrefmark{4},
Renxiang Guan \IEEEauthorrefmark{3},
Chengze Du \IEEEauthorrefmark{5},\\
Renda Han \IEEEauthorrefmark{6}, 
Xi Xuan \IEEEauthorrefmark{7},
Cuicui Luo \IEEEauthorrefmark{1}\IEEEauthorrefmark{8}} 

\IEEEauthorblockA{\IEEEauthorrefmark{1}University of Chinese Academy of Science, Beijing, China
}
\IEEEauthorblockA{\IEEEauthorrefmark{2}Harbin Institute of Technology, Harbin, China
}
\IEEEauthorblockA{\IEEEauthorrefmark{3}National University of Defense Technology, Changsha, China}
\IEEEauthorblockA{\IEEEauthorrefmark{4}Tsinghua University, Beijing, China}
\IEEEauthorblockA{\IEEEauthorrefmark{5}Beijing University of Posts and Telecommunications, Beijing, China}
\IEEEauthorblockA{\IEEEauthorrefmark{6}Hainan University, Hainan, China}
\IEEEauthorblockA{\IEEEauthorrefmark{7}City University of Hong Kong, Hong Kong SRA, China}
\IEEEauthorblockA{\IEEEauthorrefmark{8}Corresponding Author, luocuicui@ucas.ac.cn}
}

\maketitle  
\begin{abstract}
Time series anomaly detection is critical for system monitoring and risk identification, across various domains, such as finance and healthcare. However, for most reconstruction-based approaches, detecting anomalies remains a challenge due to the complexity of sequential patterns in time series data. On the one hand, reconstruction-based techniques are susceptible to computational deviation stemming from anomalies, which can lead to impure representations of normal sequence patterns. On the other hand, they often focus on the time-domain dependencies of time series, while ignoring the alignment of frequency information beyond the time domain. To address these challenges, we propose a novel Frequency-augmented Convolutional Transformer (FreCT). FreCT utilizes patch operations to generate contrastive views and employs an improved Transformer architecture integrated with a convolution module to capture long-term dependencies while preserving local topology information. The introduced frequency analysis based on Fourier transformation could enhance the model's ability to capture crucial characteristics beyond the time domain. To protect the training quality from anomalies and improve the robustness, FreCT deploys stop-gradient Kullback-Leibler (KL) divergence and absolute error to optimize consistency information in both time and frequency domains. Extensive experiments on four public datasets demonstrate that FreCT outperforms existing methods in identifying anomalies.
\end{abstract}
\begin{IEEEkeywords}
Time series, Anomaly detection, Contrastive learning.
\end{IEEEkeywords}
\section{Introduction}
Time series anomaly detection has been widely applied in many industrial applications, such as water treatment devices \cite{20234715086979}, aerospace \cite{20234314966783}, and sever machines \cite{20242216186638}. Time series anomaly detection aims to identify instances with rare or unusual patterns that can negatively affect system operations. Effectively detecting abnormal patterns is crucial to prevent industrial systems from serious detriments and unknown jeopardy. For instance, monitoring and identifying abnormal fluctuations in the data generated from photovoltaic systems can prevent system failures and even crashes in time. By deploying sophisticated anomaly detection technologies, potential issues in photovoltaic systems can be instantly perceived and settled, improving the efficiency and robustness of energy production.

Establishing an efficient and stable time series anomaly detection model remains an urgent problem to be solved. First, the definition of anomalies is often difficult to standardize and unify. For some periodic time series, irregular values or patterns may be considered anomalies, while for stable series, sudden changes or outliers might be treated as anomalies. Defining anomalies with high confidence requires extensive expert knowledge and experience, making it hard to achieve. Second, anomalies are often triggered by conjoint comportment of multiplex variables, instead of depending on a single variable, resulting in the complexity of inducing factors and the difficulty of anomaly detection. For example, legion data-driven industries, such as the Internet of Things, intelligent transportation systems, and smart logistics and supply chains, generate substantial quantities of protean and pluralistic time series data from diversified fundamental facilities and distributed control systems daily.

In this context, it is problematic to resolve complex time series anomaly detection problems utilizing shallow and conventional models. On the one hand, traditional time series anomaly detection, including statistical \cite{20241816020837} and classic machine learning methods \cite{20242016084328}, can only capture shallow representations or dependencies among sequences, while failing to model complex tendencies and discover latent discriminative space for anomalies. On the other hand, as the most effective learning tools, deep learning approaches \cite{20242516292013, 20231013683131, 20224313004438}, have been introduced into time series anomaly detection problems and achieved impressive performance for their powerful ability to extract latent features. However, to capture distinctive and representative embeddings of anomalies in such a nonlinear and intricate feature space, conventional deep learning methods based on supervised or semi-supervised paradigms must make full use of label information, which poses a challenge to data availability, especially for abnormal data. Furthermore, in real-world applications, the rarity of anomalies wreaks gross disproportion of available abnormal samples, which makes it harder to learn the regularity of behavior patterns of sequences via supervised learning paradigms.

Contrastive learning aims to adaptively learn complex representations by generating high-quality contrastive views from the inherent features of data to distinguish subtle differences between samples. As a powerful unsupervised learning framework, contrastive learning has exhibited illustrious application potential for time series anomaly detection. Specifically, contrastive learning can not only explore latent representations for samples from different categories by iteratively updating the parameters of models but also remain competitive against obstruction of deficient label information. In this way, the algorithms can accurately model the behavior pattern of time series by exploiting the superiority in quantities of benign samples and then perceive anomalies via output discrepancies and distribution isolation. 

However, most contrastive learning methods usually leverage reconstruction loss, such as mean square error (MSE), to train the model and identify anomalies, resulting in limited identification ability. On the one hand, reconstruction-based unsupervised learning methods rebuild the patterns of normal instances by calculating the difference between the real values and the generated values, and the instances that fail to be rebuilt are treated as anomalies. Since anomalies in time series usually appear with temporal dependencies instead of in the form of a single point, abnormal segments existing in normal instances can increase reconstruction loss immensely, amplifying the loss of reconstruction and obstructing the training of parameters. Consequently, it constrains the capacity of models to capture the key characteristics of normal patterns, resulting in difficulties in learning a pure pattern of normal sequences. 
On the other hand, these methods only focus on the characteristics alignment in the time domain to guarantee the model captures temporal dependencies of time series, while ignoring the alignment of frequency-domain information that cannot be exhibited in the time domain. Frequency-domain information has been empirically certified as adept at time series analysis and modelling tasks \cite{20243817050886,20242016075597,20230246821,20241715985375}. Modelling time series in the frequency domain can bypass complex auto-correlation of sequences in the time domain, which helps to improve the modelling ability and anomaly identification observations.

To address these challenges, we propose a novel Frequency-augmented Convolutional Transformer (FreCT) framework for identifying anomalies in time series by utilizing Transformer architecture to model sequences from both time and frequency domains. First, FreCT generates two contrastive views from different patch levels in the time domain by patch operations. Then a multi-layer encoder combining Transformer architecture with the convolution module is employed to capture long-range dependencies and fine-grained semantic information. Then, FreCT leverages Fourier transformation to acquire frequency-domain representations of sequences. Finally, FreCT introduces KL divergence to measure the discrepancy in the time domain and implements modulus operation to calculate the information deviation in the frequency domain.

The proposed FreCT resolves the challenges of time series anomaly detection in several ways. For the first challenge, instead of using MSE as a loss function, FreCT introduces KL divergence to moderate the problem that the loss is largely magnified by anomaly segments. In addition, a stop-gradient strategy is utilized to enhance the robustness of the training process. For the second challenge, FreCT integrates Fourier transformation with time-domain analysis to capture more critical characteristics of time series. By aligning the consistent information between two contrastive views, FreCT can accurately understand the frequency characteristics of behavior patterns of time series.

The contributions of this research include the following:
\begin{itemize}
\item We propose a novel FreCT framework for multivariate time series anomaly detection, which detects abnormal patterns of time series by measuring the consistency in the time domain and the frequency domain. 
\item  We introduce a patch-based Transformer integrated with convolution layers to capture dependencies in the time domain and evaluate consistency through a robust contrastive loss function based on KL divergence, reducing computational and spatial costs and overcoming the limitations of reconstruction loss.
\item Extensive comparative experiments are conducted on four public datasets with eleven baseline methods, demonstrating the effectiveness and superiority of FreCT over existing benchmarks.
\end{itemize}

\section{Related Work}
\subsection{Time Series Anomaly Detection}
Recently, research on time series anomaly detection has been developing rapidly. Various methods have been proposed for detecting anomalies in time series, which can be categorized into statistical approaches, classical machine learning approaches, and deep learning approaches\cite{20223012388180}.

Statistical methods include GARCH \cite{WOS:001161183300001}, CUSUM \cite{WOS:000897100800001}, regressive \cite{WOS:000946866600001}, ARIMA \cite{WOS:001046364000001}, and so on. For example, Yang et al. \cite{WOS:001146640600001} employ the EWMA model to detect early neurological deterioration in ischemic stroke patients and achieve impressive observations. Alzahrani et al. \cite{WOS:000977572900001} assemble EWMA, KNN, and CUSUM to solve underlying threats from botnets.

Machine learning methods consist of clustering \cite{WOS:001097160000025}, decision tree \cite{WOS:001176393100001}, SVM \cite{WOS:001246203300002}, and so on. For instance, Shi et al. \cite{WOS:001097160000025} leverage density peak clustering to understand the underlying patterns in time series data and to investigate the relationships between different data points for engineering applications. MTGFlow \cite{WOS:001245017200003} is an innovative framework based on a sophisticated approach that integrates dynamic graph and modelling with entity-aware normalizing flow for the unsupervised detection of anomalies in multivariate time series data, which leverages graph structure to grasp the interdependence between relations and deploys a normalizing flow to integrate unique representations of entities. Islam et al. \cite{WOS:001205820700001} apply three different machine learning models (decision tree, random forest, and XGBoost) to identify irregularities in time-series respiration information. Mukherjee et al. \cite{WOS:001221496300001} employ a one-class classifier based on a support vector machine for the detection of discordant signatures that may indicate potential equipment malfunctions or failures. As aforementioned, machine learning methods can model characteristic representations of time series, which helps to improve the accuracy of identification.

Deep learning methods include Transformer-based methods \cite{WOS:001148056000039}, recurrent neural networks \cite{WOS:000990314300004}, convolutional neural networks \cite{WOS:001240163800001}, graph neural networks \cite{WOS:001161549501027} and so on. Most of the research optimizes the model by minimizing reconstruction error or increasing the estimated probability for typical data, which could be affected by specific variations in the data\cite{20242016075056}. To this end, DCdetector \cite{DCdetector} leverages dual-channel contrastive learning to model the normal pattern of input sequences. Anomaly Transformer \cite{AnomalyTrans} deals with this problem by comparing series association with a prior distribution. VQRAEs \cite{20223512637893} designs a recurrent neural network with an objective function based on robust divergences to improve the robustness of performance. In addition, many studies focus on improving performance by exploring the interaction between variables. For example, GAT-DNS \cite{WOS:001124276300031} integrates a graph attention network with graph embedding for DNS multivariate time series anomaly detection. GIN \cite{wang2024anomaly} integrates novel graph learning and the high-efficiency Transformer to identify anomalies via adversarial training.

In general, deep learning methods exhibit superior effectiveness in time-series anomaly detection compared with traditional methods. Many studies focus on the reconstruction in the time domain while ignoring key frequency characteristics beyond the time domain. Besides, reconstruction-based methods are vulnerable to anomaly samples, resulting in the deficiency of robustness in the training process caused by amplified computational deviation and loss.
\subsection{Contrastive Representation Learning}
Contrastive representation learning aims to learn a discriminative latent embedding space where samples in the same categories have adjacent locations while samples from different categories stay far apart. Traditional contrastive learning generates positive and negative pairs, which helps models distinguish samples from different classes, such as \cite{20240215353803, 10.1007/978-981-99-8546-3_26}. Some studies get rid of strenuous negative samples and achieve comparable observations, such as SimSiam \cite{20231213775317, 20240915636021}, and BYOL \cite{20234314930993}. Besides, the Moco family \cite{20224613120228}, \cite{20224813185407} is also a competitive contrastive learning paradigm that leverages a momentum-based queue dictionary to capture fine-grained information about samples' size and consistency. To cope with model collapse, Lee et al. \cite{20241715986073} improve the guided stop-gradient strategy by exploiting the asymmetric architecture, and RecDCL \cite{20242216163598} investigates how to combine batch-wise contrastive learning with feature-wise contrastive learning.

Contrastive learning is an expandable and flexible learning paradigm, and many augmented contrastive methods have been developed for different issues. In this study, we treat two contrastive views as positive samples and learn to model a normal pattern of time series.
\subsection{Fourier Transform and Anomaly Detection}
The Fourier transform converts a time sequence into a set of orthogonal sine and cosine signals. Utilizing Euler's formula, this transformation can be further expressed in terms of complex numbers. In practical applications, given that data is often available in discrete form, many studies leverage discrete Fourier transform to convert time series to obtain information in frequency domain \cite{WOS:000977662600001, 20204909594962, WOS:000608609300001}. For example, CAFFN \cite{WOS:001157531800001} leverages a series and feature mixing block to learn representations in 1D space and utilizes a fast Fourier transform to convert representations into 2D space. ATF-UAD \cite{ATF-UAD} identifies anomalies in time series according to latent representations derived from the time and frequency domains. Labaien et al. \cite{WOS:001013582000001} develop a new positional encoding based on discrete Fourier transform for anomaly diagnosis tasks.

Accordingly, the Fourier transform plays an important role in time series information mining and is beneficial to acquiring discriminative dependencies for anomaly detection. In this paper, we implement time and frequency domain analysis in order to improve the performance of multivariate anomaly detection.
\section{Problem Statement}
In this section, we formally present the definition of multivariate time series anomaly detection.

Given a multivariate time series of length $T$, we denote the dataset as $\mathbf{X}=\{\mathbf{x}_1, \cdots, \mathbf{x}_T\}$, where $\mathbf{x}_i \in \mathbf{R}^d$ denotes the features acquired from machines or sensors at a certain timestamp $i$ and $d$ is the feature dimensionality. 
The research problem can be defined as follows: given the input time series  $\mathbf{X}_{train}$ for training, we aim to predict the labels $\mathbf{Y}_{test}$ of an unknown time series $\mathbf{X}_{test}$ for testing, where $y_i \in \mathbf{Y}_{test} = \{0, 1\}$, with 1 and 0 denoting abnormal and normal samples, respectively. To achieve this goal, we design and train a deep neural network to model the normal pattern of sequence segments based on extracted knowledge, and the segments that fail to establish a normal pattern will be classified as anomalies.

\section{Methodology}
In this section, we first present an overview of the FreCT framework. Next, we describe each component of FreCT in detail. Finally, we explain the training process.
\subsection{Overview}
\begin{figure*}
    \centering    \includegraphics[width=0.9\linewidth]{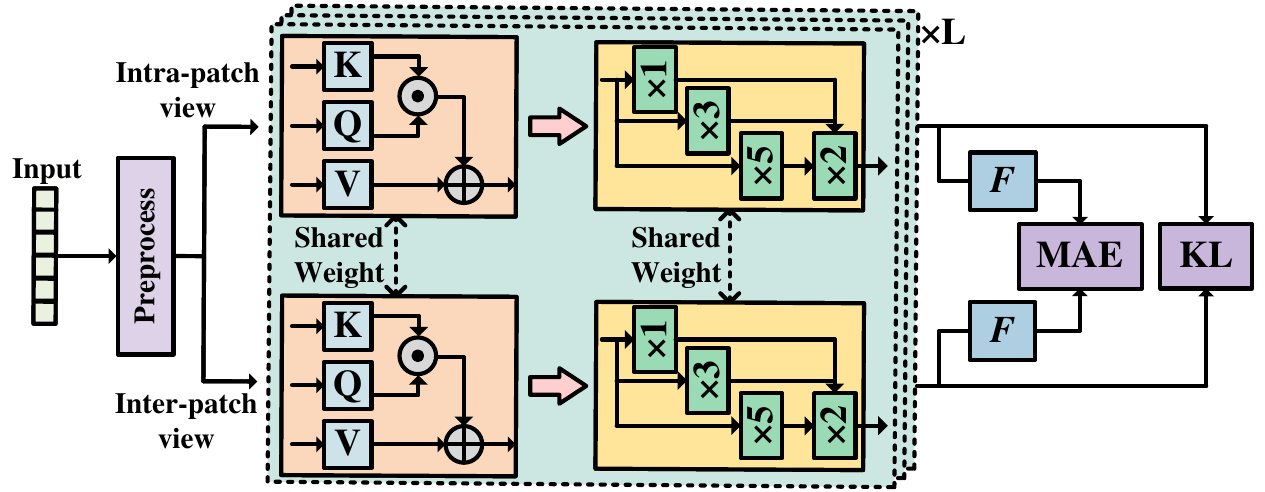}
    \caption{The overall framework of FreCT. First, FreCT leverages the preprocessing module to normalize the time series and generate patches. Then, FreCT captures intra-patch and inter-patch dependencies through the Transformer integrated with the convolution module. Then, FreCT utilizes KL-based contrastive learning to capture the consistency in the time domain and implements the Fast Fourier Transform to capture consistency in the frequency domain. Last, FreCT detects time series anomalies based on the consistencies in the time and frequency domain.}
    \label{framework}
\end{figure*}

The framework of FreCT is illustrated in Figure \ref{framework}. The input sequences first go through a sequence-level preprocess module, which includes normalization and patch generation. Sequence-level normalization improves the stability of the sequences and helps capture latent dependencies more effectively. Patch generation divides the sequence into smaller patches to enhance the representation of local semantic information. Following preprocessing, the sequences are encoded through multiple layers, each consisting of a Transformer module and a convolution module.  Transformer modules encode the long-term dependencies of time series from the perspective of inter-patch and intra-patch. The convolution module enhances the fine-grained local details of dependencies for better extracting the interrelation along the sequences. Subsequently, the encoded embeddings are transformed into frequency signals using Fourier transform methods, which improves the consistency in representing the time series. Finally, we use KL divergence and absolute error to calculate the consistency loss in both the time and frequency domains, respectively.
\subsection{Sequence-level preprocessing}
Sequence-level preprocessing mainly contains two modules, namely normalization and patch-based channel generation.

\begin{figure}[!htb]
\centering
\subfloat[The normalization module]{
		\includegraphics[scale=0.9]{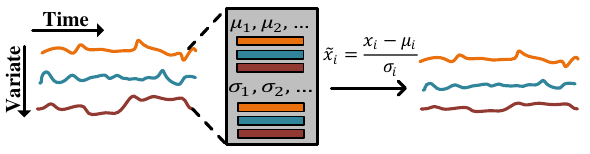} \label{normalization}}
\hfill
\subfloat[The patch-based channels generation]{
		\includegraphics[scale=0.9]{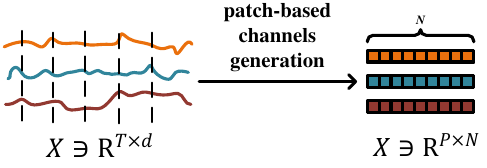} \label{patch}}
\caption{The sequence-level preprocessing, including sequence-level normalization module and patch-based channels generation.}
\label{preprosceesing}
\end{figure}

Normalizing input data helps to accelerate convergence and improve the stability of the deep learning process. Most Transformer-based frameworks employ normalization for each timestamp, standardizing the interrelationship between variables at a certain timestamp. According to \cite{itransformer}, when the event changes at that time point, the normalization toward timestamps will not only generate interactive uncertainty and interference in delayed and non-causal processes but also may cause over-smooth problems. Normalization towards sequences can deal with non-stationary issues, and its effectiveness has been successfully verified \cite{liu2022non}. In addition, sequence-level normalization can lessen the inconsistent discrepancies in the range between variables while preserving the interrelationship within sequences. Different from the majority of previous works, we implement normalization towards sequences as shown in Figure \ref{normalization}, and the formulation is as follows:
\begin{equation}
\label{normalization equation}
\mathrm{Norm}(\mathbf{V})=\{ \frac{\mathbf{v}_{i} - \mathrm{Mean}(\mathbf{v}_{i})}{\sqrt{\mathrm{Var}(\mathbf{v}_{i})}} \| i=1, \cdots, d \},
\end{equation}
where $\mathrm{Mean}(\cdot)$ and $\mathrm{Var}$ respectively denote the average and variance function, and $\mathbf{v}_{i}$ denotes the $i$-th variable.

To generate different contrastive views and capture latent dependency information from different views, we split the input sequences into different small patches, which is an effective method for Transformer architecture based on self-attention to simplify dependency representation learning. As shown in Figure \ref{patch}, each time sequence $\mathbf{X}^{i} \in \mathbb{R}^{T\times 1}(i=1, \cdots, d)$ can be divided into a series of patches and each patch can be defined as $\mathbf{X}^{i}_{n}=\{ \mathbf{x}^{i}_{1}, \cdots, \mathbf{x}^{i}_{P} \}(n=1, \cdots, N)$, where $P$ and $N$ respectively represent patch sizes and the number of patches. Generally, multivariate input sequences $\mathbf{X} \in \mathbb{R}^{T\times d}$ can be patched as $\mathbf{X} \in \mathbb{R}^{P\times N \times d}$. Then we assemble the batch dimension with feature dimensionality and the input sequence can be viewed as $\mathbf{X} \in \mathbb{R}^{P\times N}$ for simplicity. 

Note that the use of patch operations can reduce the computational complexity of Transformers, improve the quality of long-term dependency modelling, and strengthen local feature extraction through multi-scale pathways.
\subsection{Patch-based Transformer}
To model the normal pattern of time series, we implement contrastive representation learning based on Transformer architecture. Different from traditional contrastive learning methods that treat original data and augmented data as contrastive views, we leverage patch operation to generate contrastive views, which reduces the time and space complexity required to generate enhanced data, while avoiding the deviation caused by enhanced data. Specifically, we partition the time series into a series of patches and regard the inter-patch level and intra-patch level as two different contrastive views. In this way, we can avoid corrupting and changing the intrinsic information in time series, preserving the original key characteristics in time series. Besides, because there is no need to generate negative samples, this strategy saves computational costs and moderates spatial complexity.

For the inter-patch view, each patch is considered a discrete unit, enabling a structured approach to analyze its distinct characteristics, and the relationships and dependencies between these patches are effectively modelled using a multi-head self-attention mechanism. Consider the embeddings of inter-patch level time series $\mathbf{X}_{N} \in \mathbb{R}^{N \times d_{model}}$, we adopt a self-attention mechanism to encode the input sequences. Firstly, for each head, the query, key, and value matrices are initialized as follows:
\begin{equation}
\label{inter-patch}
   \mathbf{Q}_{N_{h}} = \mathbf{W}_{\mathbf{Q}_{h}} \mathbf{X}_{N_{h}},  \mathbf{K}_{N_{h}} = \mathbf{W}_{\mathbf{K}_{h}} \mathbf{X}_{N_{h}},   \mathbf{V}_{N_{h}} = \mathbf{W}_{\mathbf{V}_{h}} \mathbf{X},
\end{equation}
where $\mathbf{Q}_{N_{h}}, \mathbf{K}_{N_{h}}\in \mathbb{R}^{N \times \frac{d_{model}}{H}}$ respectively represent query, key, $\mathbf{V}_{N_{h}}\in \mathbb{R}^{P\times N\times \frac{d_{model}}{H}}$ represents value, $W_{\mathbf{Q}_{h}}, W_{\mathbf{K}_{h}}, W_{\mathbf{V}_{h}}\in \mathbb{R}^{\frac{d_{model}}{H}\times \frac{d_{model}}{H}}$ denote learnable weight matrices of $\mathbf{Q}_{N_{h}}, \mathbf{K}_{N_{h}}, \mathbf{V}_{N_{h}}$, and $h=1, \cdots, H$ and $H$ is the head number. Then the attention weights $Att_{N_{h}} \in \mathbb{R}^{N \times N}$ of different patches can be calculated as follows:
\begin{equation}
\label{inter-patch softmax}
   \mathrm{Att}_{N_{h}} = \sigma(\frac{\mathbf{Q}_{N_{h}} \cdot \mathbf{K}_{N_{h}}'}{\sqrt{\frac{d_{model}}{H}}}),
\end{equation}
where $\sigma( \cdot)$ is softmax normalization function. Note that as patching operations gain benefits from local information, the inter-patch level ignores the relevance of elements within a single patch. In other words, $\mathrm{Att}_{N_{h}} \in \mathbb{R}^{N \times N}$ is incompatible with $\mathbf{V}_{N_{h}}\in\mathbb{R}^{P\times N \times \frac{d_{model}}{H}}$ for multiplication. Therefore, to calculate Eq. \ref{inter-patch value}, an up-sampling operation should be done to the attention weights, as shown in Eq. \ref{inter-patch upsampling}:
\begin{equation}
\label{inter-patch upsampling}
   \hat{\mathrm{Att}_{N_{h}}} = \mathrm{Upsampling}(\mathrm{Att}_{N_{h}}),
\end{equation}
where $\mathrm{Upsampling}(\cdot)$ denotes the up-sampling function and $\hat{Att_{N_{h}}} \in \mathbb{R}^{P\times N \times P\times N}$ denotes the extended attention weights of the inter-patch level. Next, the sequences encoded by the self-attention mechanism can be obtained as follows:
\begin{equation}
\label{inter-patch value}
   \mathbf{Z}_{N_{h}} = \rho(\hat{\mathrm{Att}_{N_{h}}} \cdot \mathbf{V}_{N_{h}}),
\end{equation}
where $\rho$ is sigmoid activate function, and $\mathbf{Z}_{N_{h}} \in \mathbb{R}^{P\times N \times \frac{d_{model}}{H}}$ denotes the output embeddings of patch-based Transformer. Last, we concatenate multi-head representations:
\begin{equation}
\label{inter-patch cat}
   \mathbf{Z}_{N} = ||_{h=1}^{H}(\mathbf{Z}_{N_{h}}),
\end{equation}
where $||(\cdot)$ denotes concatenation function, and $\mathbf{Z}_{N}\in \mathbb{R}^{P\times N \times d_{model}}$ denotes the final inter-patch representation.

Similarly, for the intra-patch view, the self-attention mechanism is applied in patch number $N$, and the weight matrices are shared in both the inter-patch view and inter-patch view. Specifically, consider the embeddings of intra-patch level time series $\mathbf{X}_{P} \in \mathbb{R}^{P \times d_{model}}$. Firstly, for each head, the query, key, and value matrices are initialized as follows:
\begin{equation}
\label{intra-patch}
   \mathbf{Q}_{P_{h}} = \mathbf{W}_{\mathbf{Q}_{h}} \mathbf{X}_{P_{h}},  \mathbf{K}_{P_{h}} = \mathbf{W}_{\mathbf{K}_{h}} \mathbf{X}_{P_{h}},   \mathbf{V}_{P_{h}} = \mathbf{W}_{\mathbf{V}_{h}} \mathbf{X},
\end{equation}
where $\mathbf{Q}_{P_{h}}, \mathbf{K}_{P_{h}}\in \mathbb{R}^{P \times \frac{d_{model}}{H}}$ respectively represent query, key, $\mathbf{V}_{P_{h}}\in \mathbb{R}^{N \times P \times \frac{d_{model}}{H}}$ represents value, $W_{\mathbf{Q}_{h}}, W_{\mathbf{K}_{h}}, W_{\mathbf{V}_{h}}\in \mathbb{R}^{\frac{d_{model}}{H}\times \frac{d_{model}}{H}}$ denote learnable weight matrices of $\mathbf{Q}_{P_{h}}, \mathbf{K}_{P_{h}}, \mathbf{V}_{P_{h}}$. Then the attention weights can be calculated as follows:
\begin{equation}
\label{intra-patch softmax}
   \mathrm{Att}_{P_{h}} = \sigma(\frac{\mathbf{Q}_{P_{h}} \cdot \mathbf{K}_{P_{h}}'}{\sqrt{\frac{d_{model}}{H}}}),
\end{equation}
where $\mathrm{Att}_{P_{h}} \in \mathbb{R}^{P \times P}$ is the attention of different elements in patches. Since the intra-patch level neglects the relevance among patches, like Eq. \ref{inter-patch upsampling}, we extend the intra-patch attention weights:
\begin{equation}
\label{intra-patch upsampling}
   \hat{\mathrm{Att}_{P_{h}}} = \mathrm{Upsampling}(\mathrm{Att}_{P_{h}}),
\end{equation}
where $\hat{\mathrm{Att}_{P_{h}}}\in \mathbb{R}^{P\times N \times P\times N}$ denotes the extended attention weights of intra-patch level. Next the sequences encoded by the self-attention mechanism can be obtained as follows:
\begin{equation}
\label{intra-patch value}
   \mathbf{Z}_{P_{h}} = \rho(\hat{\mathrm{Att}_{P_{h}}} \cdot \mathbf{V}_{P_{h}}),
\end{equation}
where $\mathbf{Z}_{P_{h}}\in \mathbb{R}^{P\times N \times \frac{d_{model}}{H}}$ denotes the output embeddings of patch-based Transformer. Last, we concatenate multi-head representations:
\begin{equation}
\label{intra-patch cat}
   \mathbf{Z}_{P} = ||_{h=1}^{H}(\mathbf{Z}_{P_{h}}),
\end{equation}
where $\mathbf{Z}_{P}\in \mathbb{R}^{P\times N \times d_{model}}$ denotes the final intra-patch representation.
\subsection{Convolution module}
One of the most significant characteristics of the occurrence of anomalies in time series is localization, which puts forward higher requirements for the algorithm to capture the local representation. Transformer-based methods have great capability to model long-range feature information. However, in the dependency modeling process, anomalies contain plenty of local fine-grained characteristics, which can be inevitably weakened by long-range feature information, resulting in relatively limited performance. Although patch operation segments the time series into relatively local representations, the upsampling strategy may cover some important characteristics in different patches. Besides, the topological and geometric information is hard to capture by Transformers.

The primary characteristics of convolutional neural networks (CNN) are translation invariance and locality, which is the weakness of the Transformer architecture \cite{yuan2021incorporating}. Translation invariance is related to sharing weight, which makes CNN more capable of extracting structural information about local neighbors. Locality suggests the correlations between neighboring time points. Empirical research has shown the excellent performance of the integration of the Transformer and CNN \cite{duan2023dynamic, hou2024conv2former}. Therefore, we design a parallel convolution module to enhance the extraction of fine-grained semantics and model local topology in the time domain.

Specifically, consider the representation $\mathbf{Z}$ encoded by a patch-based Transformer from a single contrastive view, and for simplicity, we omit subscripts. We leverage multi-scale parallel convolution operations to extract more high-order information, which can be formulated as follows:
\begin{align}
\label{multi conv}
   &\nonumber \mathbf{Z}_{1} = \mathrm{Conv}_{1}(\mathbf{Z})\\
   &\nonumber \mathbf{Z}_{3} = \mathrm{Conv}_{3}(\mathbf{Z})\\
   &\mathbf{Z}_{5} = \mathrm{Conv}_{5}(\mathbf{Z})
\end{align}
where $\mathrm{Conv}_{k}$ denotes convolutional operators with kernel size $k$. After the convolution operation, we concatenate the high-order representations:
\begin{equation}
\label{conv cat}
   \hat{\mathbf{Z}} = ||_{k}^{\{1, 3, 5\}}(\mathbf{Z}_{k}), \quad 
\end{equation}
Finally, we fuse the multi-scale latent representations $\mathbf{Z}'$ by convolution transformation:
\begin{equation}
\label{conv proj}
   \mathbf{Z}' = \rho(\mathrm{Conv}_{2}(\hat{\mathbf{Z}}))
\end{equation}

Through the convolution module, we can obtain the multi-scale latent representations $\mathbf{Z}_{N}'$ and $\mathbf{Z}_{P}'$ with rich fine-grained local information for respectively inter-patch contrastive view and intra-patch contrastive view.
\subsection{Frequency augmentation learning}
With Transformer architecture and CNN modules, the model can dynamically assess long-range dependencies and local topology information. However, beyond the time domain, there is more valuable information to extract, which can provide additional guidelines on anomaly detection. Frequency analysis, as one of the advanced time series modelling paradigms, can bypass the dependency caused by auto-correlation in the time domain \cite{FREDF}, which has great potential for modelling normal patterns of time series. To enhance the ability to model normal patterns of time series, we deploy the Fourier transform to obtain the representation in the frequency domain. Specifically, we employ the Fast Fourier Transform (FFT) algorithm to implement transformation from the time domain to the frequency domain. The Fourier transform can be formulated as:
\begin{align}
\label{fft}
   \nonumber f_{k}(t) &= \mathrm{exp}(-j(2\pi/L)kt),\\
   F_{k} &= \int_{-\infty}^{\infty}x(t)f_{k}(t)dt,
\end{align}
Consider the output of convolution modules $\mathbf{Z}_{N}'$ and $\mathbf{Z}_{P}'$, we obtain their frequency representation according to Eq. \ref{fre trans}:
\begin{equation}
\label{fre trans}
   F_{N} = \mathcal{F}(\mathbf{Z}_{N}'), \quad F_{P} = \mathcal{F}(\mathbf{Z}_{P}'),
\end{equation}
where $\mathcal{F}(\cdot)$ denotes fast Fourier transform, $F_{N}$ represents the frequency information of $\mathbf{Z}_{N}'$ and $F_{P}$ represents the frequency information of $\mathbf{Z}_{P}'$. Consequently, we can calculate the contrastive loss of their consistency information in the frequency domain, which can be formulated as follows:
\begin{equation}
\label{fre loss}
   \mathcal{L}_{fre} = \sum_{j=1}^{J}|F_{N}(j) - F_{P}(j)|,
\end{equation}
where $|(\cdot)|$ is modulus computation and $J$ denotes the number of items in frequency representation. Note that different frequency components may vary largely in magnitude, which may finally result in unstable performance. Therefore, we leverage absolute loss instead of squared loss.
\subsection{Training}
The MSE loss function is vulnerable to data with abnormal samples, which leaves a harmful influence on the training process of FreCT. To model the normal time series pattern, we utilize Kullback-Leibler (KL) divergence to assess the consistency between representations from two contrastive views. The incentive is that anomalies exhibit rarity, and the encoded embeddings of normal points from different views should maintain consistency and similarity in latent space.

Specifically, the loss function can be formulated as follows:
\begin{equation}
\label{Lintra}
    \mathcal{L}_{\mathbf{Z}_{P}'}(\mathbf{Z}_{P}', \mathbf{Z}_{N}') = \sum KL(\mathbf{Z}_{P}', \Omega(\mathbf{Z}_{N}')) + KL(\Omega(\mathbf{Z}_{N}'), \mathbf{Z}_{P}'),
\end{equation}
\begin{equation}
\label{Linter}
    \mathcal{L}_{\mathbf{Z}_{N}'}(\mathbf{Z}_{P}', \mathbf{Z}_{N}') = \sum KL(\mathbf{Z}_{N}', \Omega(\mathbf{Z}_{P}')) + KL(\Omega(\mathbf{Z}_{P}'), \mathbf{Z}_{N}')
\end{equation}
where $\mathbf{Z}_{N}', \mathbf{Z}_{P}'$ are multi-scale representations encoded by convolution module from respectively inter-patch view and intra-patch view, $KL(\cdot)$ is KL divergence function, and $\Omega(\cdot)$ denotes stop-gradient operation. Then the loss function in the time domain can be formulated as follows:
\begin{equation}
    \mathcal{L}_{tim} = \frac{\mathcal{L}_{\mathbf{Z}_{P}'} - \mathcal{L}_{\mathbf{Z}_{N}'}}{\mathrm{len}},
\end{equation}
where $\mathrm{len}$ is the number of channels of different patch sizes. 

Note that stop-gradient operations help the model get rid of trivial solutions, which has been successfully demonstrated in the performance of model training\cite{SimSiam}. In addition, unlike traditional contrastive methods, we only use positive samples to implement contrastive learning, and this strategy helps to improve efficiency without any loss in performance and alleviates computational complexity.

The overall loss function is as follows:
\begin{equation}
    \mathcal{L} = \alpha \mathcal{L}_{tim} + (1 - \alpha)\mathcal{L}_{fre},
\end{equation}
where $\alpha$ is a hyper-parameter to balance the weight between $\mathcal{L}_{tim}$ and $\mathcal{L}_{fre}$.

\subsection{Anomaly inference}
In the anomaly inference stage, the test set is input into the well-trained model, and the anomaly score can be calculated according to consistency information from two contrastive views, consisting of consistency in the time domain and consistency in the frequency domain. The anomaly score in the time domain can be formulated as follows:
\begin{equation}
\label{anomaly score time domain}
   \mathrm{Score}_{tim}(\mathbf{X}) = \sum KL(\mathbf{Z}_{P}', \Omega(\mathbf{Z}_{N}')) + KL(\Omega(\mathbf{Z}_{P}'), \mathbf{Z}_{N}').
\end{equation}
The anomaly score in the frequency domain can be defined according to Eq. \ref{fre loss}:
\begin{equation}
\label{anomaly score fre domain}
   \mathrm{Score}_{fre}(\mathbf{X}) = \sum_{j=1}^{J}|F_{N}(j) - F_{P}(j)|.
\end{equation}
Finally, the total anomaly score can be obtained as:
\begin{equation}
\label{total anomaly score}
   \mathrm{Score}(\mathbf{X}) = \alpha \mathrm{Score}_{tim}(\mathbf{X}) + (1 - \alpha)\mathrm{Score}_{fre}(\mathbf{X}),
\end{equation}
where $\alpha$ is the shared hyper-parameter parameter. 

$\mathrm{Score}(\mathbf{X})$ is a point-wise anomaly score, and a higher value means more inconsistency. To distinguish normal and abnormal samples, a hyper-parameter threshold $\rho$ is leveraged to identify whether the point is an anomaly, as shown in Eq. \ref{score}. If the calculated score exceeds the defined threshold $\rho$, the output $\mathcal{Y}$ is an anomaly.
\begin{equation}
\label{score}
    \mathbf{Y} = 
    \begin{cases}
        0, \quad \mathrm{Score}(\mathbf{X}) < \rho\\
        1, \quad \mathrm{Score}(\mathbf{X}) \geq \rho
    \end{cases}
\end{equation}
\section{Experiments}
\subsection{Experimental Setup}
The experiments are conducted based on four publicly available datasets, MSL \cite{MSLandSMAP}, SMAP \cite{MSLandSMAP}, PSM \cite{PSM}, SWaT \cite{SWaT}. the statistics of the datasets are shown in Table \ref{DATASETS}. Dimension represents the recorded data for every timestamp; training and testing denote the number of training sets and testing sets; and anomaly rate suggests the proportion of abnormal samples.

We select eleven algorithms as baselines to verify the effectiveness of FreCT, including two traditional anomaly detection methods, LOF \cite{LOF} and DAGMM \cite{DAGMM}, and seven deep learning approaches, VAE \cite{VAE}, OmniAnomaly \cite{OmniAnomaly}, TranAD \cite{TranAD}, AnomalyTrans \cite{AnomalyTrans}, DCFF-MTAD \cite{DCFF-MTAD}, MAUT \cite{MAUT}, ATF-UAD \cite{ATF-UAD}, BTAD \cite{ma2023btad}, GIN \cite{wang2024anomaly}.

In the evaluation of experimental performance, three key metrics are relied upon: precision (P) \cite{TGC_ML_ICLR}, recall (R) \cite{S2T_ML_TNNLS}, and F1-score (F1) \cite{TMac_ML_MM}. These metrics are instrumental in assessing the accuracy and completeness of the results.

\begin{table*}[!htbp]
	\centering
	\caption{Statistics of datasets}
	\label{DATASETS}
     \resizebox{0.75\textwidth}{!}{
	\begin{tabular}{l l l l l l} 
		\hline
		Dataset & Dimension & Application & Training & Testing & Anomaly(\%)\\\hline
		MSL & 55 & Space & 58317 & 73729 & 5.54\\
		SMAP & 25 & Space & 135183 & 427617 & 13.13\\
		PSM & 26 & Server & 132481 & 87841 & 27.76\\
		SWaT & 51 & Water & 495000 & 449919 & 11.98\\\hline
	\end{tabular}}
\end{table*}

\subsection{Implementation}
The experiments are conducted using PyTorch in Python 3.9.12, deploying a single NVIDIA A40 GPU, 40GB of RAM, and a 2.60GHz Xeon (R) Gold 6240 CPU.
For the baseline methods, we reproduce them according to the source code offered by the authors.

FreCT includes three encoder layers. The hidden size is set to 128, and the number of attention heads is 1. Training epochs is set to 3. Adam Optimizer is deployed, and the learning rate is set to $10^{-4}$. 
The patch size, hidden dimension, window size and epochs are respectively set to [3, 5], 64 and 90 for the MSL dataset, [3, 5, 7], 256 and 105 for the SMAP dataset, [1, 3, 5], 256 and 60 for the PSM dataset and [3, 5, 7], 128 and 105 for the SWaT dataset.

\subsection{Performance}
We first evaluate the performance of the proposed FreCT, and the results are shown in Table \ref{performance}. 
\begin{table*}[!htb]
	\centering
	\caption{Overall performances of FreCT on four public datasets compared with eleven baselines. All results are in \%. The best results are highlighted in \textbf{bold}, and the second are \underline{underlined}.}
	\label{performance}
 \resizebox{0.95\textwidth}{!}{
	\begin{tabular}{l | l l l | l l l | l l l | l l l }
		\hline
		Dataset & \multicolumn{3}{l|}{MSL} & \multicolumn{3}{l|}{SMAP} & \multicolumn{3}{l|}{SWaT} & \multicolumn{3}{l}{PSM}\\\hline
            Metrix &P & R & F1 & P & R & F1 & P & R & F1 & P & R & F1\\\hline
            LOF(2000) & 47.72 & 85.25 & 61.18 & 58.93 & 56.33 & 57.60 & 72.15 & 65.43 & 68.62 & 57.89 & 90.49 & 70.61\\
            VAE(2012) & 72.12 & 97.12 & 82.71 & 52.39 & 59.07 & 55.53 & 49.29 & 44.95 & 47.02 & 76.09 & 92.45 & 83.48\\
            DAGMM(2018) & 89.60 & 63.63 & 74.62 & 86.45 & 56.73 & 68.51 & 8992 & 57.84 & 70.40 & 93.49 & 70.03 & 80.08\\
            OmniAnomaly(2019) & 89.02 & 86.37 & 87.67 & 92.49 & 81.99 & 86.92 & 81.42 & 84.30 & 82.83 & 88.39 & 74.46 & 80.83\\
            TranAD(2022) & 90.38 & 95.78 & 93.04 & 80.43 & 99.99 & 89.15 & 97.60 & 69.97 & 81.51 & 89.51 & 89.07 & 89.29\\
            AnomalyTrans(2022) & 91.92 & 96.03 & 93.93 & 93.59 & 99.41 & $\underline{96.41}$ & 89.10 & 99.28 & $\underline{94.22}$ & 96.14 & 95.31 & $\underline{95.72}$\\
            DCFF-MTAD(2023) & 92.57 & 94.78 & 93.66 & 97.67 & 82.68 & 89.55 & 89.56 & 91.55 & 90.56 & 93.52 & 90.17 & 91.81\\
            MAUT(2023) & 93.99 & 94.52 & $\underline{94.25}$ & 96.12 & 95.36 & 95.74 & 96.13 & 81.43 & 88.17 & 95.49 & 88.58 & 91.91\\
            ATF-UAD(2023) & 91.32 & 92.56 & 91.94 & 87.50 & 41.18 & 55.99 & 99.99 & 68.79 & 81.51 & 76.74 & 93.65 & 84.36\\
            BTAD(2023) & 80.13 & 97.58 & 87.99 & 82.74 & 99.99 & 90.56 & 99.77 & 68.79 & 81.43 & 74.25 & 86.13 & 79.75\\
            GIN(2023) & 91.34 & 97.01 & 94.02 & 84.99 & 98.53 & 91.26 & 98.91 & 76.84 & 86.49 & 86.52 & 92.03 & 89.19\\\hline
            FreCT & 92.53 & 98.15 & $\mathbf{95.26}$ & 94.25 & 98.87 & $\mathbf{96.51}$ & 92.68 & 100.00 & $\mathbf{96.20}$ & 93.13 & 98.31 & $\mathbf{97.55}$\\\hline
	\end{tabular}}
\end{table*}

\begin{table*}[!htb]
	\centering
	\caption{The results of ablation experiments on stop gradient. All results are in \%. The best results are highlighted in \textbf{bold}.}
	\label{stopablation}
\resizebox{0.95\textwidth}{!}{
\begin{tabular}{l l | l l l | l l l | l l l | l l l }
    \hline
    \multicolumn{2}{l|}{Stop gradient} & \multicolumn{3}{l|}{MSL} & \multicolumn{3}{l|}{SMAP} & \multicolumn{3}{l|}{SWaT} & \multicolumn{3}{l}{PSM}\\\hline
        Intra & Inter & P & R & F1 & P & R & F1 & P & R & F1 & P & R & F1\\\hline
        \usym{1F5F4} & \usym{1F5F4} & 91.48 & 97.90 & 94.58 & 93.43 & 99.81 & 96.51 & 93.43 & 99.96 & 96.59 & 95.19 & 98.82 & 96.97\\
        \usym{1F5F8} & \usym{1F5F4} & 91.59 & 96.12 & 93.80 & 93.79 & 98.94 & 96.30 & 90.79 & 9996 & 9516 & 9518 & 9854 & 9683\\
        \usym{1F5F4} & \usym{1F5F8} & 71.13 & 99.57 & 82.99 & 93.58 & 98.69 & 96.07 & 82.29 & 99.96 & 90.27 & 79.22 & 99.63 & 88.26\\
        \usym{1F5F8} & \usym{1F5F8} & 92.53 & 98.15 & $\mathbf{95.26}$ & 94.25 & 98.87 & $\mathbf{96.51}$ & 92.68 & 100.00 & $\mathbf{96.20}$ & 93.13 & 98.31 & $\mathbf{97.55}$\\\hline
\end{tabular}}
\end{table*}
\begin{table*}[!htb]
	\centering
	\caption{The results of ablation experiments on different modules. All results are in \%. The best results are highlighted in \textbf{bold}.}
	\label{moduleablation}
\resizebox{0.95\textwidth}{!}{
	\begin{tabular}{l | l l l | l l l | l l l | l l l }
		\hline
		Datasets & \multicolumn{3}{l|}{MSL} & \multicolumn{3}{l|}{SMAP} & \multicolumn{3}{l|}{SWaT} & \multicolumn{3}{l}{PSM}\\\hline
            Variants & P & R & F1 & P & R & F1 & P & R & F1 & P & R & F1\\\hline
            $FreCT_{norm}$ & 90.15 & 96.34 & 93.14 & 91.98 & 97.43 & 94.63 & 94.21 & 93.62 & 93.91 & 93.66 & 97.18 & 95.39\\
            $FreCT_{conv}$ & 93.12 & 95.01 & 94.05 & 90.23 & 98.18 & 94.04 & 93.14 & 98.81 & 95.89 & 91.52 & 98.08 & 94.69\\
            $FreCT_{fre}$ & 92.18 & 93.06 & 92.62 & 93.16 & 98.45 & 95.73 & 91.68 & 96.86 & 94.20 & 92.87 & 98.10 & 95.41\\
            FreCT & 92.53 & 98.15 & $\mathbf{95.26}$ & 94.25 & 98.87 & $\mathbf{96.51}$ & 9268 & 100.00 & $\mathbf{96.20}$ & 93.13 & 98.31 & $\mathbf{97.55}$\\\hline
	\end{tabular}
}
\end{table*}
For traditional methods, DAGMM performs better than LOF on all four datasets due to its powerful capacity to model the distribution of time series by stacked linear modules.

Deep learning baselines perform relatively better than traditional methods. VAE optimizes the encoded representations according to variational lower bounds, resulting in better performance than traditional methods. However, VAE overlooks the temporal dependencies among variables. OmniAnomaly which utilizes RNN to learn the dependencies in time sequences solves this gap and has a promotion in performances. Classical deep neural networks can extract the trend of time series, leading to an obvious improvement in performance. However, they have limited ability to capture long-range dependencies. In contrast, Transformer-based architecture leverages a self-attention mechanism to model the complex relationship between parts of a time series. In this way, more valuable information about historical sequences would be referred to provide richer latent features for modelling dependencies of time series. Approaches, like MAUT, TranAD and AnomalyTrans, ATF-UAD, BTAD, and GIN, leverage the self-attention mechanism to understand the latent spatiotemporal dependencies in time series and manage to model the normal pattern of time series. The average F1 values for MAUT, TranAD, AnomalyTrans, ATF-UAD, BTAD, and GIN on four datasets are respectively 92.52\%, 88.25\%, 95.07\%, 78.45\%, 84.93\%, and 90.24\%, most of which achieve impressive observations. AnomalyTrans and MAUT design advanced models based on sophisticated modules to grasp the normal pattern of times series. Due to their excellent relationship extraction and exploitation, they achieve predominant performances among these approaches. However, ATF-UAD seems to have relatively poor performance because the parity sampling strategy may undermine the native characteristics of sequences, leading to a finite capacity to construct key features of time series. To investigate the correlations of multi-dimensional sequences, DCFF-MTAD relies on graph neural networks and frequency information to capture spatiotemporal information from time series, showing nontrivial and competitive observations.

FreCT performs best among most baselines, which indicates the effectiveness of FreCT. FreCT brings 1.01\%, 0.1\%, 1.98\%, and 1.83\% promotion to the F1 score on the MSL, SMAP, SWaT, and PSM datasets, respectively compared with the best-performing baselines. This can be attributed to FreCT's capacity to extract representative dependencies of sequences based on Transformer architecture and grasp local detailed semantics using convolution modules from two contrastive views, which may provide insightful guidance for sequence learning. Moreover, frequency signals offer valuable information apart from the auto-correlation of time series and help learn different representations from the frequency domain. Therefore, FreCT can understand and model a normal pattern of time series accurately. 

\subsection{Ablation Experiments}
We first validate the effectiveness of the stop gradient strategy on the loss function. Table \ref{stopablation} shows the observations. Specifically, if no stop gradient is implemented, FreCT still does not fall into a trivial solution and has competitive performance compared to baselines. When implementing stop gradient strategies for a single contrastive view, the performances on MSL and PSM have an obvious decline but remain advanced compared to some baselines. When implementing stop gradient strategies for both contrastive views, the performance is the best, which demonstrates the utility of the stop gradient strategy.

Then we design three variants, namely $FreCT_{norm}$, $FreCT_{conv}$ and $FreCT_{fre}$, which respectively represent FreCT without normalization operation, FreCT without convolution module, and FreCT without frequency augmentation learning module to validate the effectiveness of normalization operation, convolution module, and frequency augmentation learning module. The results are shown in Table \ref{moduleablation}. From the observations, FreCT achieves the best performance compared with $FreCT_{norm}$, $FreCT_{conv}$, and $FreCT_{fre}$, which demonstrates the effectiveness of components in FreCT. The average declines for $FreCT_{norm}$, $FreCT_{conv}$ and $FreCT_{fre}$ on four datasets are respectively 1.79\%, 1.26\% and 0.97\%. For $FreCT_{norm}$, the normalization operation helps to sharpen distinct characteristics in time series and make the distribution easier to learn. For $FreCT_{conv}$, the convolution module helps to extract more interrelationships along the sequences. For $FreCT_{fre}$, a frequency augmentation learning module immune to auto-correlation can capture valuable information for modelling normal patterns of sequences from the frequency domain.

\subsection{Rationality Validation}
\begin{figure*}[!htb]
\centering
 \resizebox{0.95\textwidth}{!}{
    \subfloat[The F1 metric]{
    		\includegraphics[scale=0.3]{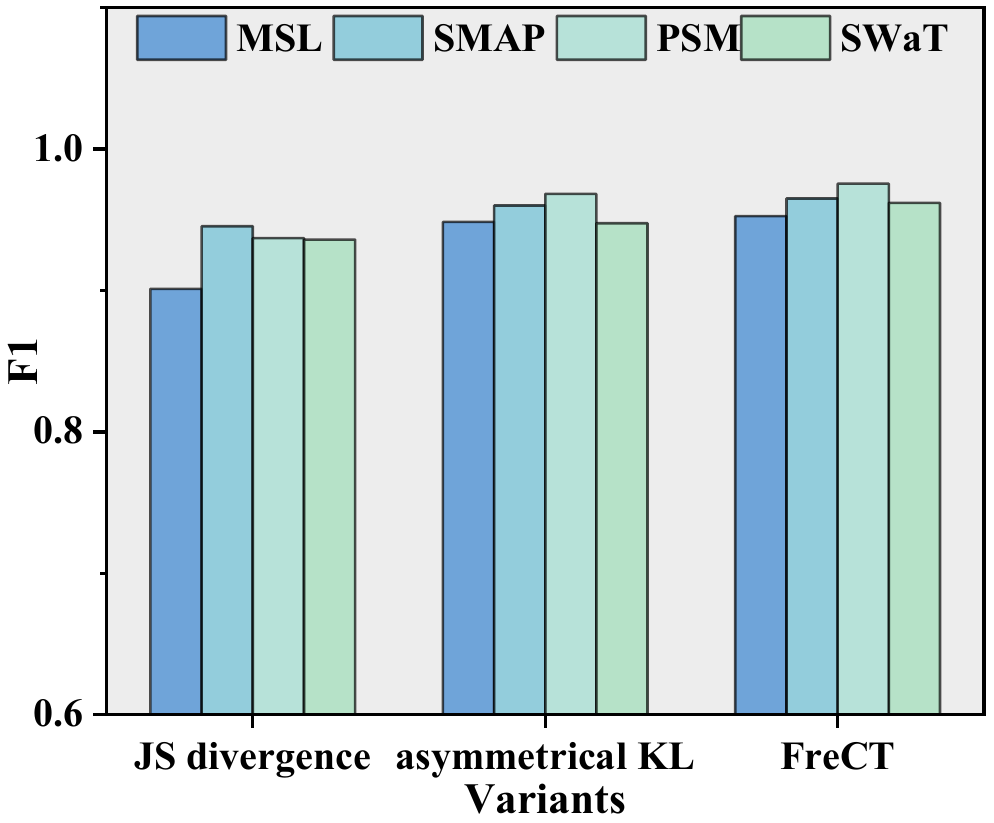} \label{F1 metric rationality validation}}
    \subfloat[The ACC metric]{
    		\includegraphics[scale=0.3]{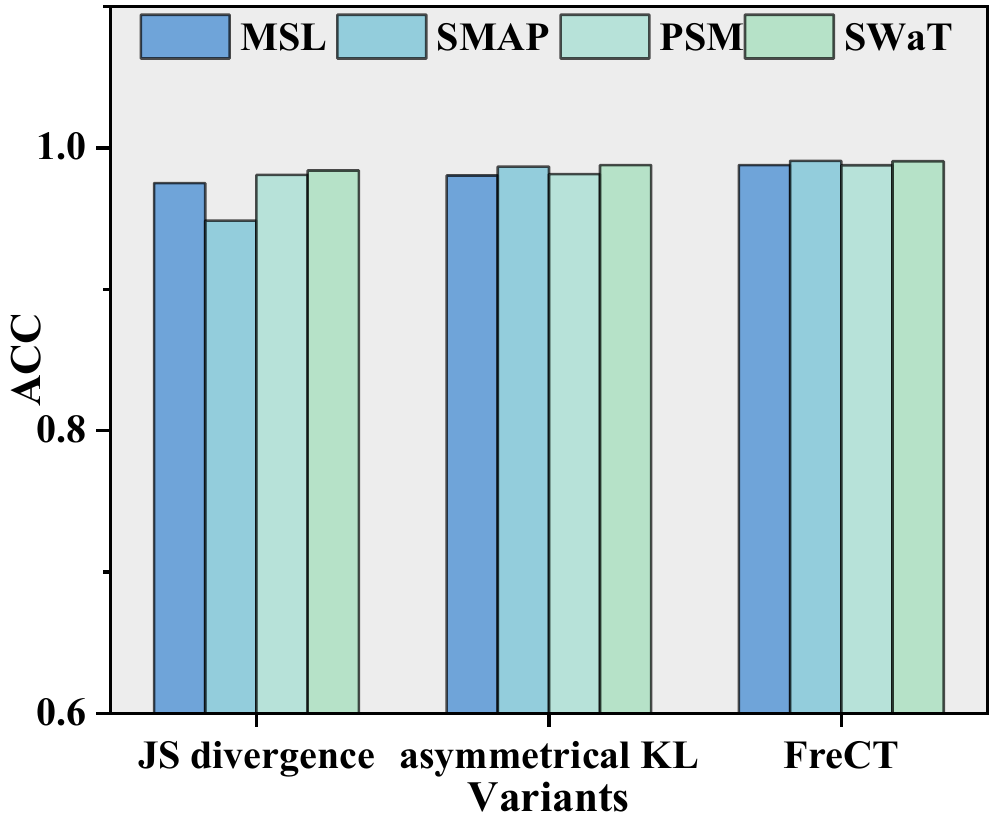} \label{ACC metric rationality validation}}
        }
\caption{The rationality validation experiments of the loss function.}
\label{rationality validation}
\end{figure*}

We first validate the rationality of the loss function. We design several variants, namely FreCT with JS divergence and FreCT with asymmetrical KL divergence, to demonstrate the effectiveness of FreCT with symmetrical KL divergence. Figure \ref{rationality validation} shows the performance of rationality validation experiments of the loss function. The F1 and ACC metrics of FreCT with three types of loss functions are respectively all over 0.8 and 0.94, which demonstrates our FreCT has a non-trivial solution according to different optimization strategies. Specifically, FreCT with JS divergence performs relatively poorly compared to FreCT with KL divergence. Furthermore, the performance of FreCT with asymmetrical KL is inferior to FreCT, which indicates symmetrical KL divergence is superior to asymmetrical KL divergence for measuring the consistency between two different distributions.

\begin{table}[!htbp]
	\centering
	\caption{The results of rational validation on frequency transform methods. All results are in \%.}
	\label{FFT validation}
    \resizebox{0.45\textwidth}{!}{
	\begin{tabular}{l| l| l l l }
		\hline
		Fourier transform & Datasets & P & R & F1 \\\hline
        \multirow{5}{*}{rfft} & MSL & 92.75 & 96.56 & 94.62\\
        ~ & SMAP & 94.50 & 97.17 & 95.82 \\
        ~ & PSM & 97.66 & 96.76 & 97.20 \\
        ~ & SWaT & 92.39 & 99.96 & 96.03 \\\hline
        \multirow{5}{*}{rfft-2d} & MSL & 92.02 & 97.48 & 94.67\\
        ~ & SMAP & 94.20 & 98.76 & 96.43 \\
        ~ & PSM & 98.10 & 92.56 & 95.25 \\
        ~ & SWaT & 91.04 & 99.96 & 95.29 \\\hline
        \multirow{5}{*}{fft} & MSL & 92.53 & 98.15 & $\mathbf{95.26}$\\
        ~ & SMAP & 94.25 & 98.87 & $\mathbf{96.51}$ \\
        ~ & PSM & 93.13 & 98.31 & $\mathbf{97.55}$ \\
        ~ & SWaT & 92.68 & 100.00 & $\mathbf{96.20}$ \\\hline
	\end{tabular}}
\end{table}
\begin{figure*}[!htb]
\centering
 \resizebox{0.95\textwidth}{!}{
    \subfloat[The F1 metric]{
    		\includegraphics[scale=0.31]{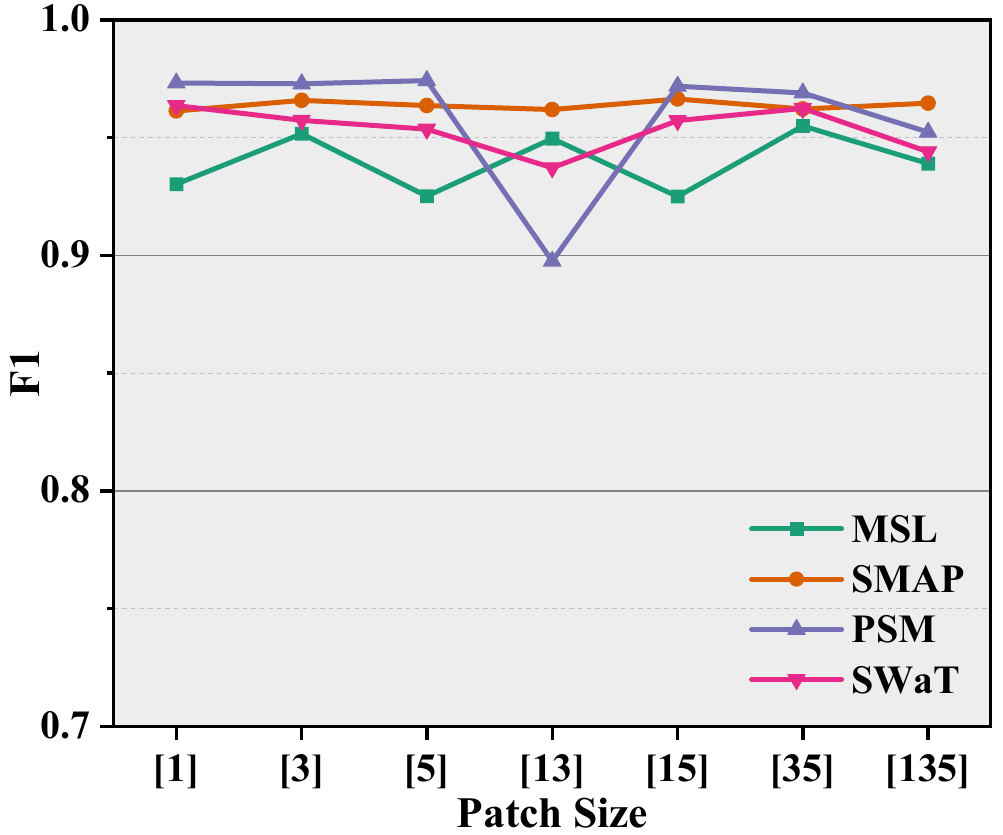} \label{F1 metric PATCH}}
    \subfloat[The ACC metric]{
    		\includegraphics[scale=0.31]{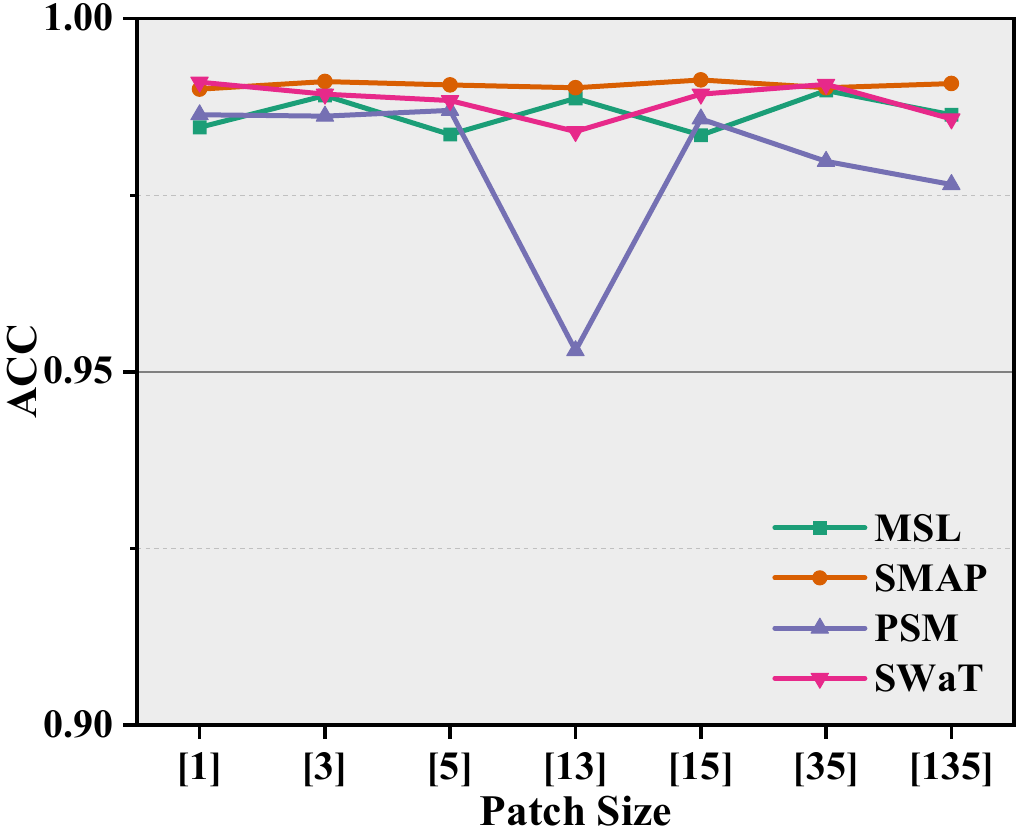} \label{ACC metric PATCH}}
        }
\caption{The sensitivity experimental results of patch size.}
\label{patch size experiments}
\end{figure*}

\begin{figure*}[!htb]
\centering
 \resizebox{0.95\textwidth}{!}{
    \subfloat[The F1 metric]{
    		\includegraphics[scale=0.31]{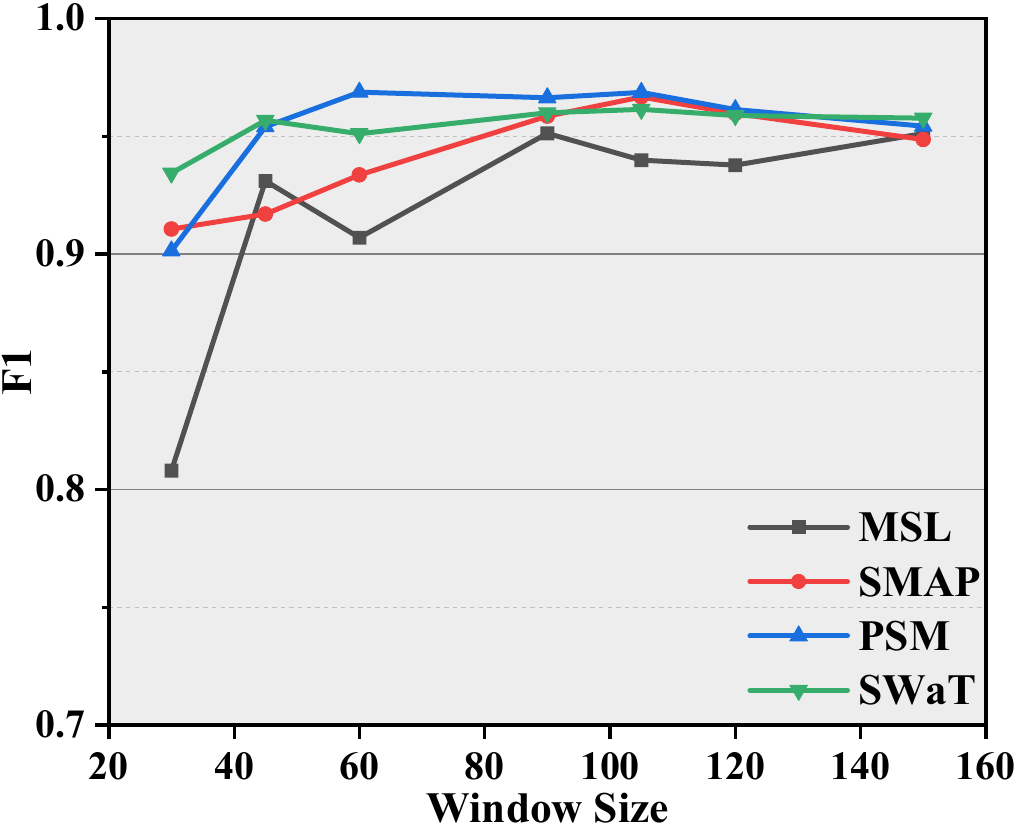} \label{F1 metric window}}
    \subfloat[The ACC metric]{
    		\includegraphics[scale=0.31]{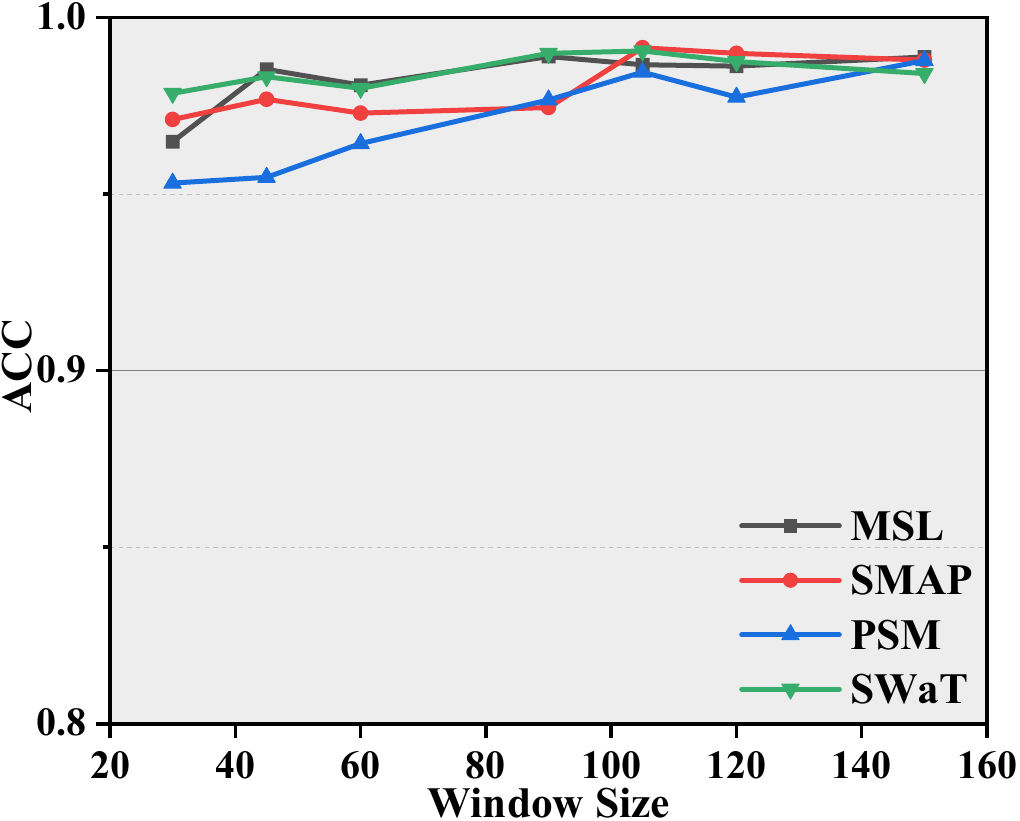} \label{ACC metric window}}
}
\caption{The sensitivity experimental results of window size.}
\label{window experiments}
\end{figure*}

\begin{figure*}[!htb]
\centering
 \resizebox{0.95\textwidth}{!}{

\subfloat[The number of layers]{
		\includegraphics[width=45mm]{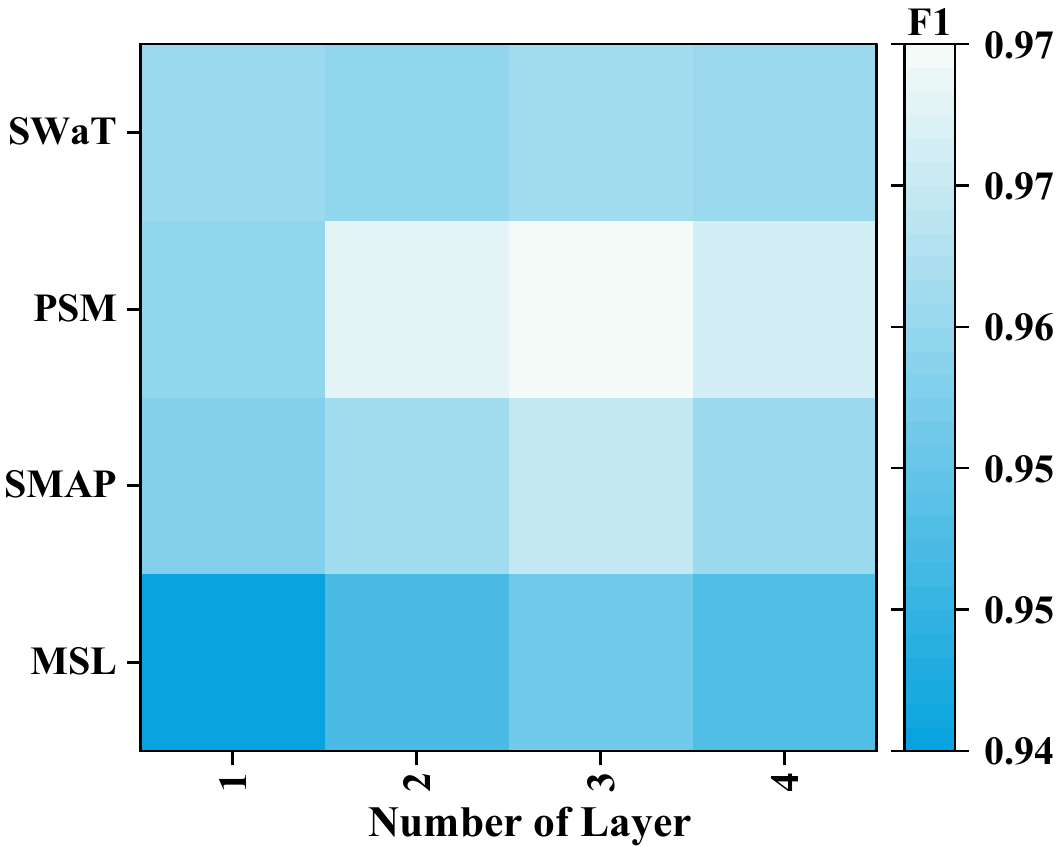} \label{F1 layers}}
\subfloat[The embeddings size]{
		\includegraphics[width=45mm]{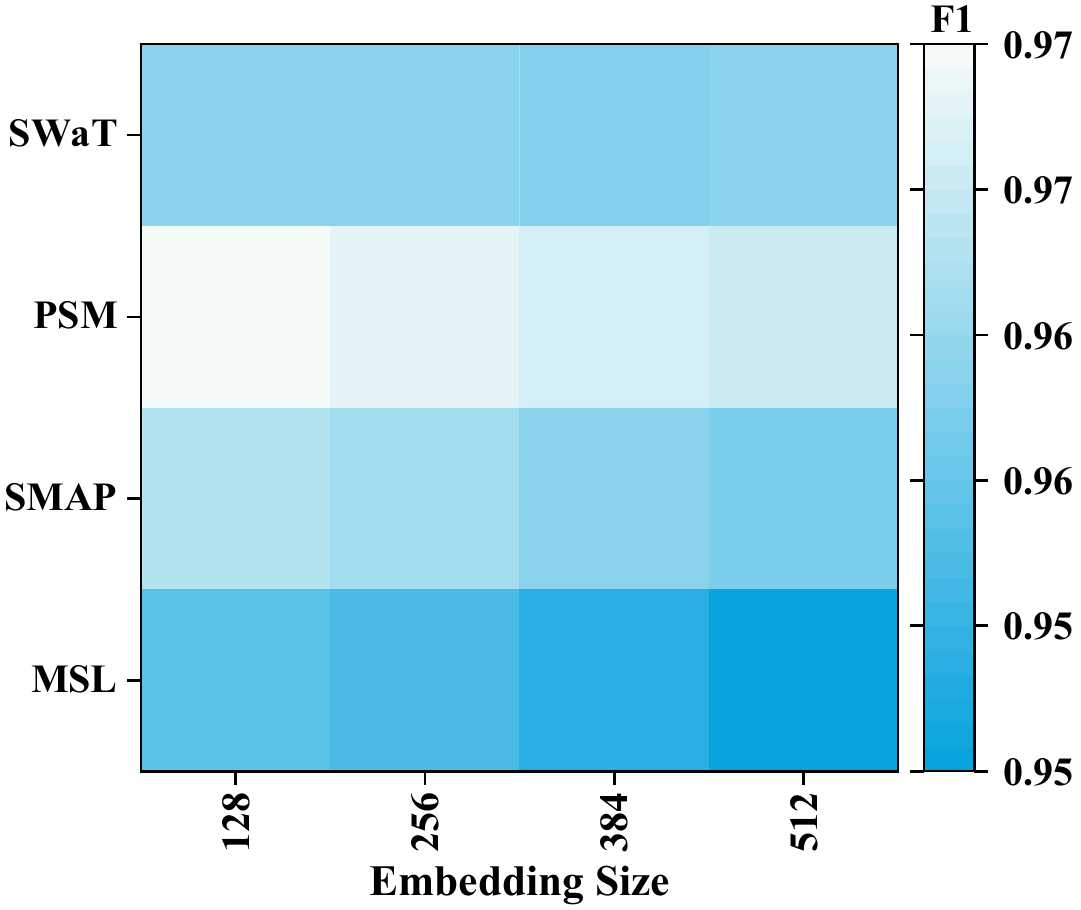} \label{F1 embeddings}}
\subfloat[The number of heads]{
		\includegraphics[width=45mm]{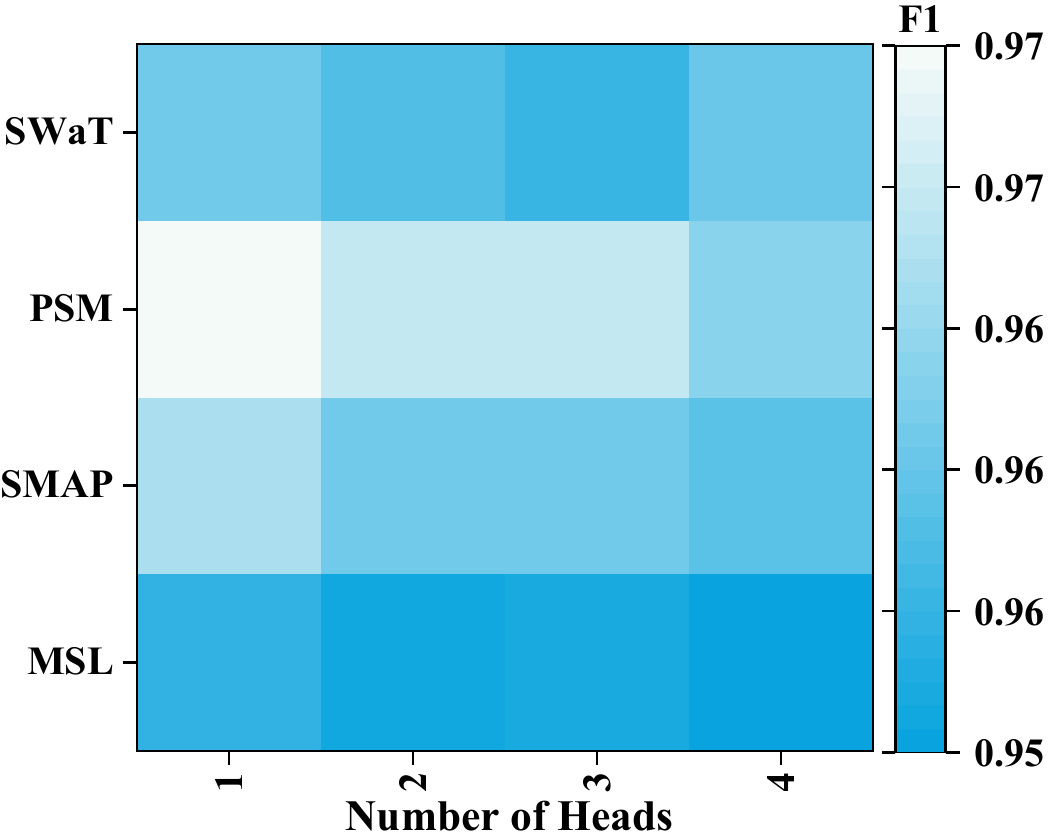} \label{F1 heads}}
}
\caption{The F1 score metric of sensitivity experiments on the number of layers, the embedding size, and the number of heads.}
\label{sensitivity experiments}
\end{figure*}
Then we verify the rationality of frequency transform methods. Transformation is implemented by mapping the time series onto a set of mutually orthogonal bases, and the selection of frequency transform methods relies on the analysis of a specific problem. We introduce several variants of the Fast Fourier transform, which analyze signals based on mutually orthogonal sinusoidal functions, and the observations are shown in Table \ref{FFT validation}, where rfft computes the one-dimensional Fourier transform of real-valued input sequences, rfft-2d computes the 2-dimensional discrete Fourier transform of real input sequences.

Through comparison of different Fourier transform methods, we can observe that fft achieves the best performance, which demonstrates the rationality of the selected frequency transform method. Note that although rfft and rfft-2d do not achieve the best performance, they still reach non-trivial solutions, indicating that our approach can extract representative features for anomaly detection problems.

\subsection{Parameter Sensitivity}
First, we test the influence of patch size on anomaly detection performance. Using different scales of patch size, multi-scale patch information can be obtained for contrastive learning. As shown in Figure \ref{patch size experiments}, the horizontal coordinate represents a set of different patch sizes. For example, [1, 3] represents that we split the time series according to patch size 1 and patch size 3. The model learns the individual representations of each patch size channel, and finally, the representations are combined. According to the observations, some conclusions can be drawn. First, the performance of FreCT varies slightly when using different patch sizes. For the PSM dataset, the performance has a little decline when the patch size shifts from [15] to [135]. Second, the optimal configuration of patch size depends on the datasets. For the dataset PSM, the best patch size is [1, 5], while for the SWaT dataset, the best choice is [3, 5]. Third, the influence of patch size varies on different datasets. For instance, when patch size changes from [5] to [13], the performance is stable on the SMAP dataset while erratic on PSM and MSL datasets.

Window size is an important hyper-parameter in time series anomaly detection that determines the length of instances. We have studies on window size in a large range [30, 150], and the observations are shown in Figure \ref{window experiments}. According to the observations, in the window size range [90, 120], the performance is relatively stable and superior. Small window sizes fail to contain sufficient sequence information, and too large window sizes may introduce interference and noise that obstruct feature learning.

In addition, we conduct sensitivity experiments on the number of layers, the embedding size, and the number of heads; the results are shown in Figure \ref{sensitivity experiments}. From Figure \ref{F1 layers}, we can conclude that when the number of layers is 3, the performance of FreCT is the best, which indicates that the number of layers also affects the ability of FreCT to learn latent representations. Figure \ref{F1 embeddings} shows the influence of embedding size on the performance of FreCT. Generally speaking, small embedding sizes may result in a deficiency in feature representations, and too large embedding sizes will cause model collapse. From Figure \ref{F1 heads}, we can see that the best performance is achieved with the number of heads being 1. The multi-head attention mechanism will make the representations insufficient for distinction.

\subsection{Efficiency}
\begin{figure}[!htb]
    \centering
    \includegraphics[width=0.55\linewidth]{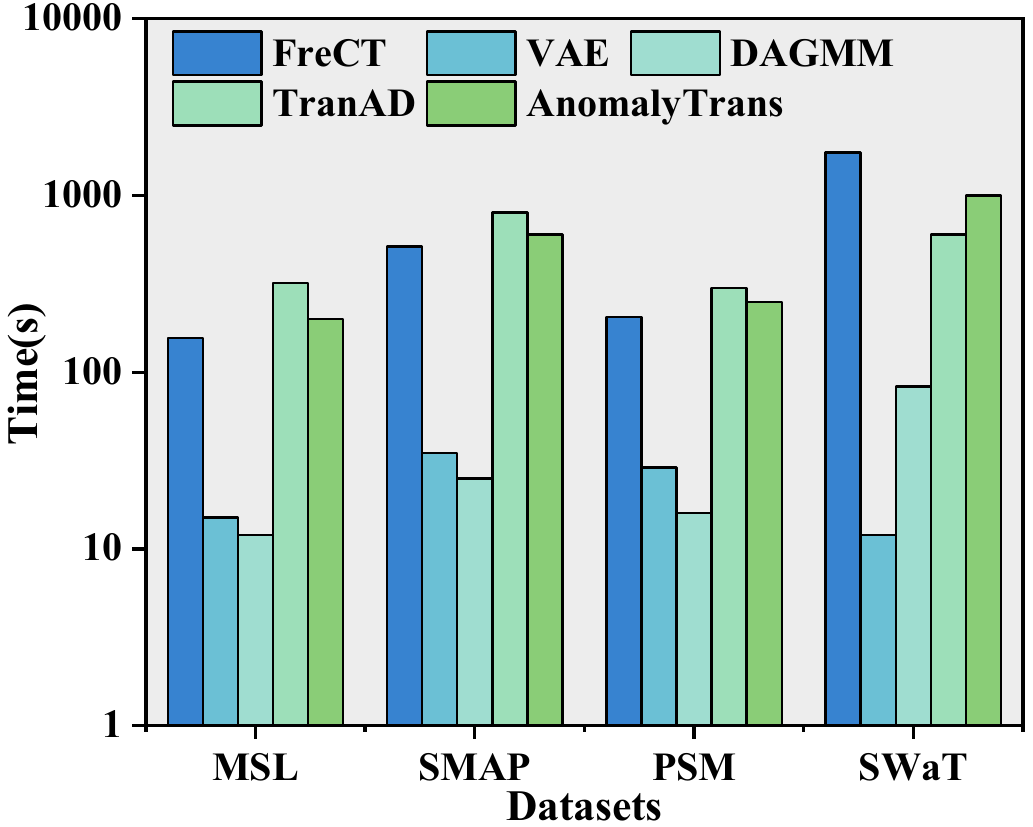}
    \caption{The efficiency observation of FreCT compared with several baselines.}
    \label{time}
\end{figure}

Our subject is to detect anomalies in time series. Efficient and timely detection can help reduce the loss caused by anomalies. To assess the efficiency of FreCT, we compare the training time of FreCT with several selected state-of-the-art algorithms, and the observations are shown in Figure \ref{time}. We choose traditional machine learning and statistical methods and some relatively superior methods instead of all baselines, which can be regarded as the upper bounds of all the baselines. Therefore, the selected algorithms are representative for comparison with FreCT in the time efficiency experiment, which can illustrate the efficiency performance of FreCT.

From the observations, the required time of FreCT for training is shorter than that of TranAD and AnomalyTrans on most datasets, showing that dual-channel contrastive consistency information learning is more efficient than association discrepancy fitting learning and dual auto-encoder architecture. Besides, leveraging CNN to extract latent embeddings makes the features more representative and more discriminative, boosting the identification efficiency, while the training of FreCT costs more time than DAGMM and VAE. The first reason is that DAGMM is a traditional method that does not have many parameters to learn, and VAE is a lightweight foundational model that has fewer parameters than state-of-the-art algorithms. The second reason is that they have limited ability to learn latent and discriminative representations, so they leverage less time to train.

To sum up, our proposed FreCT not only has great and competitive performance but also has relatively superior efficiency.

\section{Conclusion}
Time series anomaly detection is a critical mission for system monitoring. An effective and efficient anomaly detection method can improve the performance of the system. This study introduces a novel approach called Frequency-augmented Convolutional Transformer (FreCT) for anomaly detection. FreCT employs unsupervised contrastive learning to analyze data from both the time domain and the frequency domain, achieving strong performance. This approach integrates convolution with Transformers to enhance long-range dependencies and local semantic details and leverages Fourier transforms to capture frequency information. Extensive experiments are conducted to demonstrate the superiority of FreCT compared to various state-of-the-art baselines.

In the future, we are committed to addressing the temporal limitations of anomaly detection algorithms to ensure timely anomaly detection and alerting. Additionally, we will further investigate the computational scalability of these algorithms and their deployment in resource-constrained environments.

\section{Acknowledgment} 
This work was supported by the National Natural Science Foundation of China under Grant 72210107001, the Beijing Natural Science Foundation under Grant IS23128, the Fundamental Research Funds for the Central Universities, and by the CAS PIFI International Outstanding Team Project (2024PG0013).

%
%
%
\bibliographystyle{IEEEtran}
\bibliography{ref}

\end{document}